\documentclass[sigconf]{acmart}

\usepackage{amsmath,amsfonts}
\usepackage{algorithmic}
\usepackage{graphicx}
\usepackage{textcomp}
\usepackage{xcolor}
\usepackage{glossaries}
\usepackage{textcomp}
\usepackage[capitalise]{cleveref}
\usepackage{svg}
\usepackage[inline]{enumitem}
\usepackage{siunitx}
\usepackage{booktabs}

\newcommand{\mround}[1]{\num[round-mode=places,round-precision=0]{#1}}

\AtBeginDocument{%
  \providecommand\BibTeX{{%
    \normalfont B\kern-0.5em{\scshape i\kern-0.25em b}\kern-0.8em\TeX}}}

\newacronym{cnn}{CNN}{convolutional neural network}
\newacronym{map}{mAP}{mean average precision}
\newacronym{qat}{QAT}{quantization-aware training}
\newacronym{ptq}{PTQ}{post-training quantization}
\newacronym{gpu}{GPU}{graphics processing unit}
\newacronym{cpu}{CPU}{central processing unit}
\newacronym{fpga}{FPGA}{field programmable gate array}
\newacronym{asic}{ASIC}{application-specific integrated circuit}
\newacronym{soc}{SoC}{system on chip}
\newacronym{yolo}{YOLO}{"You Only Look Once"}
\newacronym{rcnn}{R-CNN}{region-based convolutional neural network}
\newacronym{ssd}{SSD}{single shot detector}
\newacronym{mcu}{MCU}{microcontroller unit}
\newacronym{ml}{ML}{machine learning}
\newacronym{ai}{AI}{artificial intelligence}
\newacronym{dnn}{DNN}{deep neural network}
\newacronym{bbox}{Bbox}{bounding box}
\newacronym{ilp}{ILP}{integer linear programming}
\newacronym{iot}{IoT}{Internet of Things}
\newacronym{tqt}{TQT}{trained quantization thresholds}
\newacronym{pulp}{PULP}{parallel ultra-low-power}

\setcopyright{iw3c2w3g}
\copyrightyear{2023}
\acmYear{-}

\acmConference['23]{preprint}{preprint}{preprint}
\def\eqref#1{Eq.~(\ref{#1})}

\begin{document}

\title{Flexible and  Fully Quantized Ultra-Lightweight TinyissimoYOLO for Ultra-Low-Power Edge Systems}


\author{Julian Moosmann}
\authornote{All three authors contributed equally to this research.}
\orcid{1234-5678-9012}
\affiliation{%
  \institution{ETH Zurich}
  \city{Zurich}
  \country{Switzerland}}
\email{julian.moosmann@pbl.ee.ethz.ch}

\author{Hanna Müller}
\authornotemark[1]
\affiliation{%
  \institution{ETH Zurich}
  \city{Zurich}
  \country{Switzerland}}
\email{hanna.mueller@iis.ee.ethz.ch}

\author{Nicky Zimmerman}
\authornotemark[1]
\affiliation{%
  \institution{University of Lugano}
  \city{Lugano}
  \country{Switzerland}}
\email{nicky.zimmerman@idsia.ch}

\author{Georg Rutishauser}
\affiliation{%
  \institution{ETH Zurich}
  \city{Zurich}
  \country{Switzerland}}
\email{georg.rutishauser@iis.ee.ethz.ch}

\author{Luca Benini}
\affiliation{%
  \institution{ETH Zurich/University of Bologna}
  \city{Zurich/Bologna}
  \country{Switzerland/Italy}}
\email{luca.benini@iis.ee.ethz.ch}

\author{Michele Magno}
\affiliation{%
  \institution{ETH Zurich}
  \city{Zurich}
  \country{Switzerland}}
\email{michele.magno@pbl.ee.ethz.ch}

\hyphenation{Tiny-issimo-YOLO}


\begin{abstract}
  This paper deploys and explores variants of TinyissimoYOLO, a highly flexible and fully quantized ultra-lightweight object detection network designed for edge systems with a power envelope of a few milliwatts. With experimental measurements, we present a comprehensive characterization of the network's detection performance, exploring the impact of various parameters, including input resolution, number of object classes, and hidden layer adjustments. We deploy variants of TinyissimoYOLO  on state-of-the-art ultra-low-power extreme edge platforms, presenting an in-depth  a comparison on latency, energy efficiency, and their ability to efficiently parallelize the workload. In particular, the paper presents a comparison between a novel RISC-V-based parallel processor (GAP9 from Greenwaves) with and without use of its on-chip hardware accelerator, an ARM Cortex-M7 core (STM32H7 from ST Microelectronics), two ARM Cortex-M4 cores (STM32L4 from STM and Apollo4b from Ambiq), and a multi-core platform aimed at edge AI applications with a CNN hardware accelerator (Analog Devices MAX78000). Experimental results show that the GAP9's hardware accelerator achieves the lowest inference latency and energy at \SI{2.12}{\ms} and \SI{150}{\micro\joule} respectively, which is around 2x faster and 20\% more efficient than the next best platform, the MAX78000. The hardware accelerator of GAP9 can even run an increased resolution version of TinyissimoYOLO with $112\times112$ pixels and 10 detection classes within \SI{3.2}{ms}, consuming \SI{245}{\micro\joule}. To showcase the competitiveness of a versatile general-purpose system we also deployed and profiled a multi-core implementation on GAP9 at different operating points, achieving \SI{11.3}{\ms} with the lowest-latency and \SI{490}{\micro\joule} with the most energy-efficient configuration.  With this paper, we demonstrate the flexibility of TinyissimoYOLO and prove its detection accuracy detection datasets. We demonstrate its suitability for real-time ultra-low-power edge inference by benchmarking its performance.
\end{abstract}




\begin{teaserfigure}
\centerline{\includegraphics[width=0.9\textwidth]{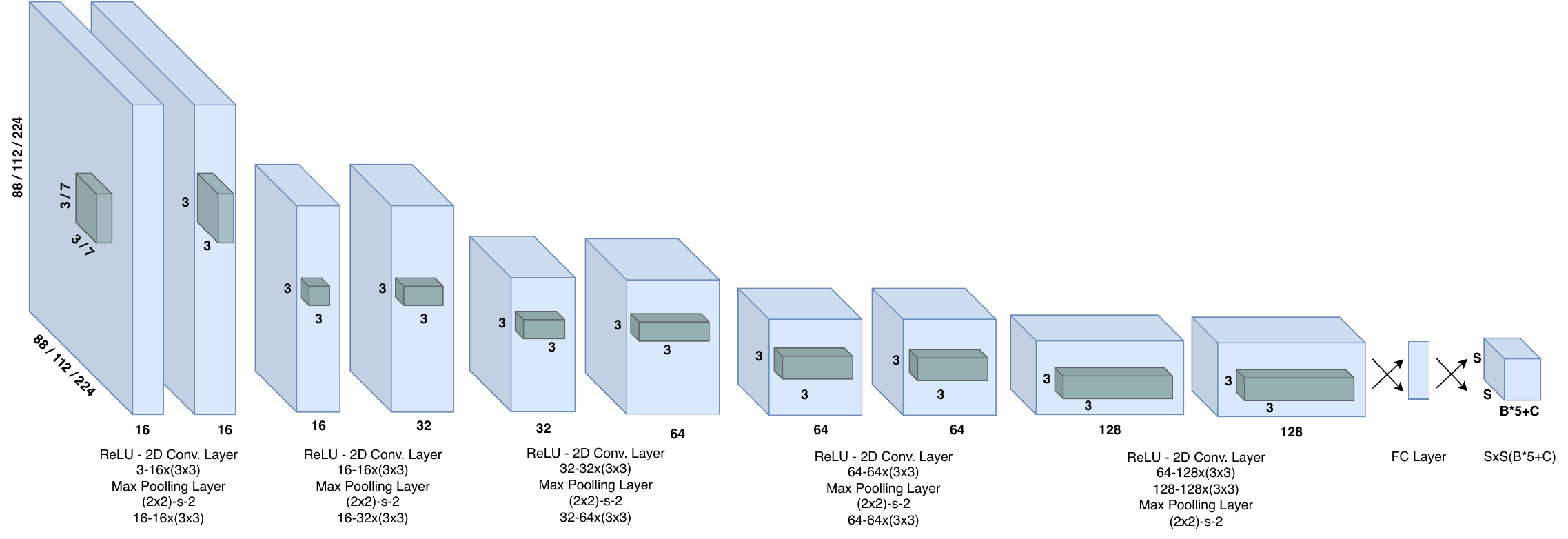}}
\caption{TinyissimoYOLO as deployed on GAP9 (both the 8-core cluster and the hardware accelerator) by this paper.}
\label{fig:tinyissimoYOLO}
\end{teaserfigure}


\maketitle

\section{Introduction}
With the widespread adoption of edge devices, particularly \gls{mcu} based nodes, the \gls{iot} is revolutionizing various domains including health monitoring, smart manufacturing, and home automation~\cite{statistaIoT}. These low-power devices enable increased automation, cost reduction, bandwidth optimization, and enhanced privacy by processing collected data \emph{on the edge}, i.e., directly on the sensor node. However, the limited memory and computing resources of \glspl{mcu} pose a significant challenge when it comes to deploying advanced machine learning models. 
Overcoming this challenge and enabling efficient machine learning on resource-constrained devices is a crucial area of research in embedded networked sensor systems, as illustrated by~\cite{ling2022blastnet}.

A foundational element of many \gls{iot} applications is the extraction of semantic information about the environment with image sensors. Specifically, object detection~\cite{Carion2020eccv, liu2016eccv}, the vital task of identifying and precisely localizing objects within a given image, plays a pivotal role in a wide range of systems. For instance, in the realm of autonomous mobile agents, object detection enables obstacle avoidance and tracking, path planning, and scene understanding, contributing to safe and efficient navigation. Similarly, in augmented and virtual reality devices, such as head-mounted displays, lightweight and energy-efficient object detection algorithms are crucial to enable real-time visual recognition without exceeding the limited power resources of the wearable device.  
A majority of recent research on object detection focuses on perfecting detection accuracy. State-of-the-art models have on the order of $10^8$ or more parameters, requiring power-hungry hardware such as GPUs~\cite{Wang_2023_CVPR, dagli2023cppe5} for inference. Therefore, state-of-the-art models like the \gls{yolo} series of networks~\cite{redmon_you_2016,redmon2017yolo9000,redmon2018yolov3,bochkovskiy2020yolov4,Jocher_YOLOv5_by_Ultralytics_2020,li2022yolov6,wang2023yolov7,Jocher_YOLO_by_Ultralytics_2023} cannot be directly ported to low power edge processors due to the memory and compute constraints of \glspl{mcu}.
Consequently, there is a growing demand for approaches to enable semantic understanding on edge ultra-low power, constrained hardware. This shift has led to a surge of interest in various research areas, including architecture search, quantization techniques, and advanced inference engines tailored for resource-constrained devices~\cite{saha2022sensors, warden2019oreilly, nni2021github, moosmann2023tinyissimoyolo}. \glspl{mcu} are now being equipped with novel open-source energy-efficient cores, such as RISC-V cores, parallel processing engines, dedicated hardware accelerators, and specialized co-processors aimed at enabling efficient execution of complex machine learning tasks~\cite{zimmerman2022iot, lamberti2021iscas}.



By combining these advancements in parallel processing, hardware accelerators, and quantization techniques, \glspl{mcu} are now executing quite sophisticated \gls{ml} models. 
However, deploying machine learning on \glspl{mcu} is still far from trivial and mapping networks like \gls{yolo} for advanced visual tasks, beyond simple classification, is still at the boundary of feasibility.

This work presents a cutting-edge \gls{yolo}-based network designed to push the boundaries and to achieve accurate and real-time object detection on \glspl{mcu} at sub-millijoule inference energy consumption. By leveraging novel low-power processors, we move from TinyissimoYOLO~\cite{moosmann2023tinyissimoyolo} and present an improved, highly flexible network architecture -- flexible TinyissimoYOLO, shown in \cref{fig:tinyissimoYOLO}. Our approach capitalizes on the unique capabilities of the latest generation of \gls{mcu} platforms, harnessing their computational capabilities to set a new state of the art in energy-efficiency object detection, with a memory footprint of below \SI{500}{\kilo\byte}. Through extensive evaluation and performance analysis, we demonstrate the potential of our network in enabling energy-efficient computer vision applications. This research contributes to the advancement of energy-efficient processing on resource-constrained devices, opening up new possibilities for a wide range of real-world applications. To support our contribution, we evaluate a complete pipeline for deploying accurate, lightweight quantized object detection on several novel \glspl{mcu}, bridging the gap between state-of-the-art models and on-edge execution. We provide a detailed description for each part of our pipeline, relying as much as possible on open-source tools, with the hope of enabling others to successfully deploy object detection models on ultra-low-power edge systems.

To quantify the benefits of the proposed approach, we evaluate the performance and energy efficiency of our object detection pipeline running on several \glspl{mcu}, executing different TinyissimoYOLO networks on one or multiple general-purpose cores as well as specialized \gls{cnn} accelerators. 
We investigate and expand the TinyissimoYOLO network  proposed by Moosmann et al.~\cite{moosmann2023tinyissimoyolo}, a lightweight general multi-object detection network optimized for a single processor, by extending its flexibility. Depending on the target platform's capabilities, TinyissimoYOLO can be scaled to make the best use of the target hardware and offer optimal performance. We deploy TinyissimoYOLO on multiple microcontroller architectures: the GAP9 multi-core RISC-V \gls{mcu} (Greenwaves), which features a hardware accelerator for \glspl{cnn}, MAX78000 (Analog Devices), a multi-core platform with an energy efficient \gls{cnn} accelerator, Apollo4b (Ambiq), the most power-efficient ARM Cortex-M4 core currently available, and STM32L4R9 and STM32H7A3 from STMicroelectronics to establish a baseline with the most popular ARM cortex-M4F and ARM cortex-M7 cores. 

Furthermore, for GAP9, we map a Pareto front of operating points (supply voltage and operating frequency) to evaluate the trade-off between latency and inference energy. In addition, we evaluate the per-layer inference power consumption and quantify the efficiency enhancements enabled by the integrated NE16 \gls{cnn} accelerator. 

 Additionally, we explore several variations of TinyissimoYOLO, highlighting the trade-offs between prediction accuracy and resource consumption. We evaluate the detection performance for a set of image resolutions and different kernel sizes applied in the first layer. We report the size of the network for each modification and the detection accuracy. Our investigation demonstrates that, as expected, increased input resolution contributes to higher detection accuracy, but also emphasized the increasing memory consumption.

The rest of this paper is organized as follows: \Cref{sec:relwork} provides an overview of the works focusing on \gls{cnn} optimization and deployment on microcontrollers. \Cref{sec:background} details our deployment pipeline, from choosing the architecture and training, to quantization and porting to target hardware. \Cref{sec:results} specifies the experimental setup and reports the results, presenting an in-depth analysis of the trade-off between performance and energy efficiency. Furthermore, we compare our GAP9-deployed model against state-of-the-art deployments on different \glspl{mcu}, focusing on latency, energy efficiency and inference efficiency (MACs/cycle). Lastly, \Cref{sec:conclusion} concludes our work.

\section{Related Work}
\label{sec:relwork}
In the past decade, deep learning approaches have revolutionized the field of image-based scene understanding, through object detection \cite{Carion2020eccv, liu2016eccv}and semantic segmentation \cite{he2017iccv, long2015cvpr}. Semantic understanding of objects in the environment is an essential capability for autonomous agents, for tasks such as localization \cite{zimmerman2023ral}, mapping \cite{zimmerman2023iros, mccormac20183dv, rosinol2020icra}, and navigation \cite{crespo2020mdpi}.  You Only Look Once or commonly abbreviated as YOLO is one of the most popular and optimized deep learning algorithms used to perform real-time detection \cite{jiang2022review}. To effectively detect and track objects, YOLO uses a repurposed classifier or localization which is a model applied to an image at several locations and scales \cite{jiang2022review}. However, the majority of works require power-hungry hardware such as GPUs and are not suitable for deployment on low-power edge devices. Even some of the more resource-conscious approaches (\cite{jiang2022review, seichter2021icra}) still require powerful hardware consuming multiple watts and requires several tens megabytes of memory to run inference at sensor-rate. Based on the recent literature, we observed that existing YOLO approaches tend to have high memory requirements, limiting their applicability on resource-constrained devices. To address this limitation, in this paper, we evaluate and optimize a novel and flexible lightweight algorithm inspired by YOLOv1. The algorithm is specifically designed to achieve optimal performance in terms of accuracy while keeping the memory requirements below \SI{500}{\kilo\byte}. This breakthrough in memory efficiency makes our algorithm well-suited for low-power \glspl{mcu}, which serve as the processing units for a wide range of extreme edge devices. By significantly reducing memory demands without compromising accuracy, our approach opens new avenues for deploying efficient and accurate object detection on resource-constrained devices.


The deployment of image classification and object detection models on \glspl{mcu} has garnered significant attention in recent years. Canepa et al.~\cite{canepa2022icecs} propose a method for detecting specific objects in surveillance video frames using deep neural networks on an STM32 \glspl{mcu}. Although they achieve high prediction accuracy, their slow inference rate (\SI{0.03}{\hertz}) limits its suitability for real-time applications, and their power consumption of approximately \SI{400}{\mW} is relatively high. To enable semantic understanding on edge devices and small autonomous agents, the models must be small and efficient enough to be executed on low-memory (<\SI{1}{\mega\byte}) and ultra-low-power platforms (<\SI{100}{\mW}) at significantly higher inference rates. 

This task requires a specialized workflow, composed of three main strategies. 
Firstly, exploiting the benefits of resource-aware neural architecture search \cite{fedorov2019nips, risso2023toc, liberis2021arxiv, benmeziane2021IJCAI}. Secondly, pruning and quantization strategies \cite{han2015arxiv, liang2021neurocomp}, as offered by commonly used deep learning frameworks such as PyTorch. Notable quantization frameworks targeting ultra-low-power hardware include TensorFlow Lite \cite{warden2019oreilly} and  Microsoft NNI \cite{nni2021github}, and other academic platforms. 
And lastly, deployment on lightweight hardware using inference engines, which aim to improve data locality, memory usage, and spatiotemporal execution. 

TinyML software suites\cite{saha2022sensors}, including the open-sourced TensorFlow Lite Micro\cite{david2021mlsys}, EdgeML\cite{kumar2017icml} and CMSIS-NN \cite{lai2018arxiv}, allow for deploying neural networks on \glspl{mcu} and are mainly designed for ARM Cortex-M and as such less attractive for RISC-V based processors.
Similarly, Wulfert et al.\cite{wulfert2022icfsp} present an object detection method for resource-limited systems, performing camera-based human detection directly on a small ESP32 \glspl{mcu}. While they achieve a high inference rate of \SI{12}{\hertz}, their approach is limited to detecting a single class. Likewise, Palossi et al.\cite{zimmerman2022iot} demonstrate real-time human tracking on a nano drone, mounted with GAP8, a RISC-V parallel platform from Greenwaves, but are constrained to a single class. Lamberti et al.~\cite{lamberti2021iscas} propose a specialized low-power Automatic License Plate Recognition system executed on GAP8 at an approximate frequency of \SI{1}{\hertz}. 

In contrast to these prior works, our paper makes several novel contributions targeting a variety of promising platforms. Firstly, our proposed flexible lightweight algorithm ensures outstanding performance in terms of accuracy while keeping the memory requirements low enough for deployment on \glspl{mcu}. Secondly, we target specifically different low-power \glspl{mcu}, which serve as the processing units for a wide range of edge devices. This diverse platform evaluation further highlights the versatility and robustness of our approach. Overall, this work addresses the limitations of existing methods and introduces a memory-efficient algorithm suitable for resource-constrained devices, opening up new possibilities for efficient and accurate object detection in real-world applications. Additionally, the evaluation of our proposed algorithm on different platforms, including ARM, RISC-V cores and hardware accelerators, provides valuable insights into the benefits and trade-offs associated with each hardware architecture. This comparative analysis allows us to identify the strengths and weaknesses of each platform, enabling us to make informed decisions based on the specific requirements of the application at hand.
This analysis provides a comprehensive view of the various approaches, highlighting the benefits of parallel processing in the multi-core RISC-V processor and the efficiency gains achieved through hardware accelerators. Furthermore, it offers a valuable perspective on power consumption, latency, and scalability, that influence the choice of hardware for object detection tasks on resource-constrained devices. 

Giordano et al.~\cite{giordano2022aicas} benchmark a single architecture for image classification on several different platforms. Moss et al.~\cite{moss2022gihthub} evaluate different image classification architectures on a single platform, MAX78000. Unlike these works, we describe the full deployment pipeline in the context of object detection, from architecture exploration to quantization and hardware-optimized implementation.

 In the realm of efficient neural architecture design for \glspl{mcu}, MCUNet~\cite{lin2020nips} presents a framework that combines the lightweight inference engine (TinyEngine) with the efficient neural architecture (TinyNAS), enabling ImageNet-scale inference on \glspl{mcu}. Building upon this, MCUNetV2~\cite{lin2021nips} introduces memory-efficient patch-based inference, further enhancing memory performance for image classification and object detection. However, while these works report peak memory consumption and MACs, they lack comprehensive power consumption measures or run-time evaluations. By exploring various variations of the TinyissimoYOLO~\cite{moosmann2023tinyissimoyolo} network and optimizing it to be deployable in different platforms, we provide a detailed analysis of the trade-offs between prediction accuracy, run-time, and power consumption across multiple deployment platforms. Notably, we demonstrate the feasibility of reducing the MACs to approximately 3 million, representing a 10-50 times reduction compared to MCUNetV2. In contrast to the previously published TinyissimoYOLO paper, which evaluated the network's performance on a restricted subset of the PascalVOC~\cite{everingham_pascal_2015} dataset and only on single-core processors, this work investigates the network's performance without any restrictions on the dataset and explore and benchmark the flexibility and the hardware overall energy efficiency. This allows us to evaluate the performance of our flexible TinyissimoYOLO version across the entire PascalVOC dataset, considering all classes and unrestricted object counts within each image. By combining advancements in architecture design, memory-efficient inference, quantization-aware training, and a comprehensive evaluation across various deployment platforms, our work makes significant contributions to the field. We showcase the remarkable reduction in MACs achieved by our proposed approach, surpassing the state-of-the-art MCUNetV2. Furthermore, our thorough evaluation on the complete PascalVOC dataset demonstrates the robustness and scalability of our flexible TinyissimoYOLO network. Overall, our work pushes the boundaries of energy-efficient object detection on \glspl{mcu}, providing valuable insights and paving the way for further advancements in the field.




\section{Background and Implementation}
\label{sec:background}
In this section, we describe \emph{TinyissimoYOLO}, which we used as the basis for our explorations, training, and dataset, the MCU platforms we compare to and deploy on, and the deployment tools. 
\subsection{TinyissimoYOLO}
TinyissimoYOLO is a general multi-object detection network, designed to enable fast and accurate detection on microcontroller platforms. Its architecture is shown in \Cref{fig:tinyissimoYOLO}. 
The original network uses an input resolution of $88\times 88$ pixels and produces an output vector of dimension $\left( S\times S\left( B\times 5+C\right)\right)$, where $S\times S$ is the grid of prediction cells, $B$ is the number of boxes predicted per cell and $C$ is the number of classes. This architecture is highly flexible: increasing the input resolution trades improved detection performance for increased computational load and a proportional increase in the number of parameters of the last layer. The number of detected classes $C$ can also be varied, affecting only the parameter count and computational volume in the last layer.
The \gls{cnn} backbone used for feature extraction is small in comparison to state-of-the-art object detection networks. The original \gls{yolo}v1 network has 20 GMAC and 45M parameters, while the more recent \gls{yolo}v7 scales from 1.75 GMAC to 420 GMAC and 6.2M to 151M parameters. In comparison, the deployed TinyissimoYOLO can be scaled from 32 MMAC to 57 MMAC and from 441K to 888K parameters. Using 8-bit quantization further reduces the model size and memory footprint by a factor of $4$ when compared to equivalent 32-bit floating-point models, making TinyissimoYOLO ideally suited for real-time inference on resource-constrained \gls{mcu} platforms.

In this work, we investigate the influence of network parametrization on detection performance. We train TinyissimoYOLO with different input resolutions  ($88\times 88$, $112\times112$ and $224\times 224$ pixels), different numbers of output classes $C$ ($3$, $10$ and $20$) and different kernel sizes in the first layer ($3\times 3$ and $7\times 7$).



\begin{table*}[htbp]
    \caption{TinyissimoYOLO network trained and evaluated on PascalVOC with different network configurations. This table shows the network performances for the different network configurations. The naming convention of the different network configurations is: TY(TinyissimoYOLO):classes-1st layer's kernel-input resolution, for more details see: \cref{tab:network_variation_overview}.}
    \begin{center}
    \setlength\tabcolsep{1.5pt}
    \begin{tabular}{@{}lccccccccccccccccccccc@{}}\toprule
        \textbf{Network} & \multicolumn{20}{c}{\textbf{\gls{map}}} \\
        \cmidrule{2-22}
        \textbf{Arch.} & \textbf{\textit{mAP}}& \textbf{aero} & \textbf{bike} & \textbf{bird} & \textbf{boat} & \textbf{bottle} & \textbf{bus} & \textbf{car} & \textbf{cat} & \textbf{chair} & \textbf{cow} & \textbf{table} & \textbf{dog} & \textbf{horse} & \textbf{mbike} & \textbf{person} & \textbf{plant} & \textbf{sheep} & \textbf{sofa} & \textbf{train} & \textbf{tv} \\
        \midrule
        \textbf{TY:3-3-88}& \mround{61.76823333}\% & & & & & & &\mround{70.4635}\% & & \mround{45.1178}\% & & & & & & \mround{69.7234}\% & & & & \\ 
        \textbf{TY:3-7-88}& \mround{61.5178}\% & & & & & & &\mround{70.5656}\% & & \mround{45.2369}\% & & & & & & \mround{68.7509}\% & & & & \\ 
        \textbf{TY:3-3-112}& \mround{63.0549666667}\% & & & & & & &\mround{73.432}\% & & \mround{45.708}\%  & & & & & & \mround{70.0249}\% & & & & \\ 
        \textbf{TY:3-7-112}& \mround{61.93206667}\% & & & & & & &\mround{71.6727}\% & & \mround{46.2784}\% & & & & & & \mround{67.8451}\% & & & & \\ 
        \textbf{TY:3-3-224}& \mround{53.1861666667}\% & & & & & & &\mround{60.9841}\% & & \mround{34.5441}\% & & & & & & \mround{64.0303}\% & & & & \\ 
        \textbf{TY:10-3-88}& \mround{58.26349}\% & \mround{69.5876}\% & \mround{65.5537}\% & \mround{51.1587}\% & \mround{52.3497}\% & \mround{27.69}\%   & \mround{72.0332}\% & \mround{69.1494}\% & \mround{69.2288}\% & \mround{44.3832}\% & \mround{61.5006}\% & & & & & & & & & & \\ 
        \textbf{TY:10-7-88}& \mround{58.4205}\% & \mround{68.0647}\% & \mround{67.1346}\% & \mround{54.2149}\% & \mround{53.8335}\% & \mround{27.3138}\% & \mround{69.7332}\% & \mround{68.7567}\% & \mround{69.2147}\% & \mround{45.9825}\% & \mround{59.9564}\% & & & & & & & & & & \\ 
        \textbf{TY:10-3-112}& \mround{60.42115}\% & \mround{71.0753}\% & \mround{70.7076}\% & \mround{53.9943}\% & \mround{56.8559}\% & \mround{30.0802}\% & \mround{73.3131}\% & \mround{72.6702}\% & \mround{69.3923}\% & \mround{46.7337}\% & \mround{59.3889}\% & & & & & & & & & & \\ 
        \textbf{TY:10-7-112}& \mround{56.94412}\% & \mround{68.3342}\% & \mround{66.7097}\% & \mround{49.6355}\% & \mround{50.0218}\% & \mround{21.8204}\% & \mround{70.6228}\% & \mround{69.1351}\% & \mround{67.6023}\% & \mround{45.167}\%  & \mround{60.3924}\% & & & & & & & & & & \\ 
        \textbf{TY:10-3-224}& \mround{43.58928}\% & \mround{61.4848}\% & \mround{54.1508}\%  & \mround{38.3857}\% & \mround{34.8103}\%  & \mround{0.0}\% & \mround{50.5797}\% & \mround{59.1809}\% & \mround{64.1879}\% & \mround{34.8708}\% & \mround{38.2419}\%  & & & & & & & & & & \\ 
        \textbf{TY:20-3-88}& \mround{53.149902}\% & \mround{72.4055}\% & \mround{68.4199}\% & \mround{52.8951}\% & \mround{52.1583}\% & 0\%        & \mround{70.082}\%  & \mround{71.4452}\% & \mround{70.3253}\% & \mround{38.3479}\% & \mround{44.2658}\% & \mround{64.8779}\% & \mround{66.3338}\% & \mround{73.0614}\% & \mround{66.9815}\% & \mround{65.4055}\% & \mround{05.1502}\% & \mround{54.7154}\% & \mround{08.62944}\% & \mround{69.0806}\% & \mround{48.4173}\%  \\ 
        \textbf{TY:20-7-88}& \mround{46.997535}\% & \mround{67.3714}\% & \mround{60.4756}\% & \mround{41.7426}\%  & \mround{46.1543}\%  & 0\%        & \mround{69.8844}\% & \mround{68.0628}\% & \mround{66.4157}\% & \mround{36.0372}\% & \mround{11.6279}\% & \mround{62.6474}\% & \mround{59.3321}\% & \mround{65.6017}\% & \mround{65.5251}\% & \mround{63.6294}\% & 0\%        & \mround{54.8063}\% & \mround{23.7154}\%      & \mround{64.6641}\% & \mround{49.0352}\% \\ 
        \textbf{TY:20-3-112}& \mround{56.4352}\% & \mround{70.598}\%  & \mround{69.6407}\% & \mround{53.8825}\% & \mround{53.9491}\% & \mround{04.0909}\% & \mround{71.393}\%  & \mround{72.7902}\% & \mround{68.8156}\% & \mround{39.2073}\% & \mround{38.0769}\% & \mround{64.2194}\% & \mround{66.0926}\% & \mround{73.7217}\% & \mround{68.0277}\% & \mround{66.4526}\% & \mround{23.0924}\% & \mround{61.5738}\% & \mround{36.0927}\%  & \mround{70.7464}\% & \mround{56.2405}\%  \\ 
        \textbf{TY:20-7-112}& \mround{53.479071}\% & \mround{72.3831}\%  & \mround{63.644}\% & \mround{52.9089}\% & \mround{55.4938}\% & \mround{0.0}\% & \mround{70.8838}\%  & \mround{70.6953}\% & \mround{66.7582}\% & \mround{36.5547}\% & \mround{33.9606}\% & \mround{62.8112}\% & \mround{62.018}\% & \mround{74.1779}\% & \mround{67.3428}\% & \mround{65.0371}\% & \mround{05.01012}\% & \mround{53.122}\% & \mround{35.3077}\%  & \mround{68.4195}\% & \mround{53.0527}\%  \\ 
        \textbf{TY:20-3-224}& \mround{22.5912255}\% & \mround{58.6421}\%  & \mround{09.88593}\% & \mround{0.0}\% & \mround{0.0}\% & \mround{0.0}\% & \mround{47.0893}\%  & \mround{52.0387}\% & \mround{56.2823}\% & \mround{15.8148}\% & \mround{0.0}\% & \mround{0.0}\% & \mround{38.0771}\% & \mround{07.04698}\% & \mround{32.7074}\% & \mround{55.0775}\% & \mround{0.0}\% & \mround{20.7692}\% & \mround{0.0}\%  & \mround{47.1498}\% & \mround{11.2434}\%  \\ 
        \hline
        \textbf{YOLOv1} & &   &   &   &   &   &   &   &   &   &   &   &   &   &   &   &   &   &   &   &  \\
        \textbf{20-7-7x7-448} & \textit{58\%} & 77\%  & 67\%  & 58\%  & 38\%  & 23\%  & 68\%  & 56\%  & 81\%  & 36\%  & 61\%  & 49\%  & 77\%  & 72\%  & 71\%  & 64\%  & 29\%  & 52\%  & 55\%  & 74\%  & 51\% \\
        \bottomrule
        \end{tabular}
    \label{tab:map_tinyissimoyolo}
    \end{center}
\end{table*}

\begin{table}[htbp]
    \caption{TinyissimoYOLO network trained and evaluated on PascalVOC. This table shows the different network configurations which are evaluated. Additionally, the number of network parameters and the corresponding network model size is shown. }
    \begin{center}
    \begin{tabular}{@{}lcccrr@{}}\toprule
        & & \textbf{1st} & & & \\
        \textbf{Network} & \textbf{\#} & \textbf{layer's} & \textbf{input} & \textbf{net.} & \textbf{model}\\
         & \textbf{classes} & \textbf{kernel} & \textbf{res.} & \textbf{param.} & \textbf{Size}\\
        \midrule
        \textbf{TY:3-3-88}      & 3 & 3 & $88\times88$ & \num[group-separator={,}]{440592} & \SI{441}{\kibi\byte} \\
        \textbf{TY:3-7-88}      & 3 & 7 & $88\times88$ & \num[group-separator={,}]{442512} & \SI{443}{\kibi\byte} \\
        \textbf{TY:3-3-112}     & 3 & 3 & $112\times112$ & \num[group-separator={,}]{573712} & \SI{574}{\kibi\byte} \\
        \textbf{TY:3-7-112}     & 3 & 7 & $112\times112$ & \num[group-separator={,}]{575632} & \SI{576}{\kibi\byte} \\
        \textbf{TY:3-3-224}     & 3 & 3 & $224\times224$ & \num[group-separator={,}]{1638672} & \SI{1.64}{\mebi\byte} \\
        \textbf{TY:3-7-224}     & 3 & 7 & $224\times224$ & \num[group-separator={,}]{1657872} & \SI{1.66}{\mebi\byte} \\
        \textbf{TY:10-3-88}     & 10 & 3 & $88\times88$ & \num[group-separator={,}]{498048} & \SI{498}{\kibi\byte} \\
        \textbf{TY:10-7-88}     & 10 & 7 & $88\times88$ & \num[group-separator={,}]{499968} & \SI{500}{\kibi\byte} \\
        \textbf{TY:10-3-112}    & 10 & 3 & $112\times112$ & \num[group-separator={,}]{702848} & \SI{703}{\kibi\byte} \\
        \textbf{TY:10-7-112}    & 10 & 7 & $112\times112$ & \num[group-separator={,}]{722048} & \SI{722}{\kibi\byte} \\
        \textbf{TY:10-3-224}    & 10 & 3 & $224\times224$ & \num[group-separator={,}]{2341248} & \SI{2.34}{\mebi\byte} \\
        \textbf{TY:10-7-224}    & 10 & 7 & $224\times224$ & \num[group-separator={,}]{2360448} & \SI{2.36}{\mebi\byte} \\
        \textbf{TY:20-3-88}     & 20 & 3 & $88\times88$ & \num[group-separator={,}]{580128} & \SI{580}{\kibi\byte} \\
        \textbf{TY:20-7-88}     & 20 & 7 & $88\times88$ & \num[group-separator={,}]{582048} & \SI{582}{\kibi\byte} \\
        \textbf{TY:20-3-112}    & 20 & 3 & $112\times112$ & \num[group-separator={,}]{887328} & \SI{887}{\kibi\byte} \\
        \textbf{TY:20-7-112}    & 20 & 7 & $112\times112$ & \num[group-separator={,}]{906528} & \SI{907}{\kibi\byte} \\
        \textbf{TY:20-3-224}    & 20 & 3 & $224\times224$& \num[group-separator={,}]{3344928} & \SI{3.34}{\mebi\byte} \\
        \textbf{TY:20-7-224}    & 20 & 7 & $224\times224$ & \num[group-separator={,}]{3346848} & \SI{3.35}{\mebi\byte} \\
        \bottomrule
        \end{tabular}
    \label{tab:network_variation_overview}
    \end{center}
\end{table}

\subsection{Training and Dataset}
For training, testing and validation of the TinyissimoYOLO variants, we used the PascalVOC dataset~\cite{everingham_pascal_2015}. $\SI{90}{\percent}$ of the PascalVOC training dataset was used to train the network, with the remaining $\SI{10}{\percent}$ serving as the validation set. The training data was augmented with geometric operations such as cropping, scaling, and shifting, as well as photometric operations including blurring, and modifying the brightness, contrast, saturation, and hue.

 We used the SGD \cite{saad_-line_1999} optimizer with the cosine annealing scheduler proposed in \cite{loshchilov2017arxiv}.  The hyperparameters used depend on the input size and the chosen number of classes, and are detailed for the GAP9-deployed networks in \cref{tab:hyperparameter_overview}.

 \begin{table}[htbp]
    \caption{Hyperparameter overview when training TinyissimoYOLO in QuantLab using a learning rate scheduler (cosine annealing).}
    \begin{center}
    \begin{tabular}{@{}lrr@{}}\toprule
        \textbf{Hyperparameter} & \textbf{Value} & \textbf{Note}\\
        \midrule
        \multicolumn{3}{l}{\textbf{Training}} \\
         \textit{Batch Size} & 64 & \\
         $LR_0$ & $10^{-3}$ & $224\times224$ input: \\ 
         $LR$ min & $10^{-6}$ &  \\
         \textit{Total Epochs} & 1500 & \\
         \midrule
         \midrule
         \multicolumn{3}{l}{\textbf{QAT}} \\
         \textit{QAT Epochs} & 1000-1500 & Weights quant. at 1000,  \\
          & & activations at 1200 \\ 
         \textit{\acrshort{qat} Algo.} & \acrshort{tqt} & \\
        \bottomrule
        \end{tabular}
    \label{tab:hyperparameter_overview}
    \end{center}
\end{table}




\subsection{Network Quantization}
We used \emph{QuantLab}\footnote{\url{https://github.com/pulp-platform/quantlab}} to train and quantize the networks deployed to GAP9's cluster. QuantLab is a modular, PyTorch-based framework for \gls{qat}, offering experiment management facilities, support for various quantization algorithms, and automated model conversion functionality from full-precision to fake-quantized and fully integerized models. We quantized TinyissimoYOLO networks to 8-bit weight and activation precision using the \gls{tqt} algorithm~\cite{jain2020trained} \cref{tab:hyperparameter_overview}. A QuantLab experiment starts from a standard full-precision PyTorch network, which is converted to its trainable, fake-quantized version according to a user-defined configuration. The quantized network is then trained according to the experiment specification. Finally, the trained fake-quantized model can be automatically converted to an integer-only model. In the integerized model, normalization, rescaling and activation layers are merged into requantization layers. A requantization layer consists of channel-wise integer multiplication, channel-wise addition, logical right shift and clipping, effectively executing an affine transformation and clipping in fixed-point arithmetic. This approach has been described multiple times in literature and has been variously termed "integer channel normalization"~\cite{ref:rusci_mixed} or "dyadic quantization"~\cite{ref:hawqv3}. Finally, the integerized model is exported as a backend- and hardware-agnostic ONNX model, where the exported ONNX operators are annotated with precision information, allowing the deployment backend to select the correct kernels. Both the fake-quantized TinyissimoYOLO models trained in QuantLab and the deployable integer-only models generated from them exhibited no accuracy drop compared to their full-precision counterparts.

For the models mapped to the MAX78000 platform, we use the same training and quantization procedures as in~\cite{moosmann2023tinyissimoyolo}. \\ Since the only purpose of deploying the network on the two ARM Cortex-M4 and the ARM Cortex-M7 is to assess the power and performance metrics of those single-core processors running the network inference, the network was implemented in TensorFlow, trained for a few epochs and quantized with TensorFlow-Lite. Similarly, the model deployed to GAP9's NE16 accelerator was quantized with the \gls{ptq} flow integrated in Greenwaves' NNTool, as no NE16 deployment backend for QuantLab-integerized models was available at the time of our experiments. We did not evaluate the resulting model's accuracy, as the principal purpose of the NE16 deployment was to evaluate the hardware's performance and efficiency. However, it is important to note that NE16 is fully compatible with the integerized models produced by QuantLab, including the quantized TinyissimoYOLO models whose performance we report in \Cref{tab:map_tinyissimoyolo}.

\subsection{MCU Platforms}
We compare deployments on different \gls{mcu} platforms, which we introduce here.
\subsubsection{ARM Cortex-M4 and Cortex-M7}
The used \glspl{mcu} from STMicroelectronics (STM32H7A3 and STM32L4R9) each feature an ARM Cortex-M single-core processor. The H7A3 and L4R9 \glspl{mcu} use a Cortex-M7 and a Cortex-M4 operating at up to 280MHz and 120MHz respectively, with core voltages of up to 1.3V.

Apollo4b also uses an ARM Cortex-M4 processor running up to 192MHz with a nominal core voltage of 0.65V. Apollo4b's main distinguishing characteristic is its utilization of Ambiq's proprietary subthreshold power-optimized technology platform, designed to offer maximum power efficiency for edge applications. 

MAX78000 features an ARM Cortex-M4, a built-in \gls{cnn} accelerator which has 64 specialized processors with built-in convolutional engine, pooling unit and dedicated \SI{442}{\kilo\byte} weight memory. In addition, a 32-bit RISC-V coprocessor supports ultra-low-power signal processing.

\subsubsection{GAP9}
GAP9 features 10 RISC-V cores. One core acts as a \emph{fabric controller}, orchestrating system operation, while a \gls{pulp} cluster of 9 cores implementing custom instruction set extensions is available for efficient, high-performance execution of compute-intensive tasks. Additionally, it includes NE16, a dedicated on-chip hardware accelerator for \gls{cnn} inference. The GAP9 architecture is based on the open-source \gls{soc} Vega~\cite{rossi2021vega}. The cores' maximum operating frequency is \SI{370}{\mega\hertz} for both the 9-core cluster and the fabric controller. For additional flexibility, GAP9 is provisioned for dynamic voltage and frequency scaling, allowing users to trade-off between latency and energy efficiency. The compute cluster, consisting of 9 cores, one for orchestration and 8 workers, offers general-purpose compute power at extreme energy efficiency while the \gls{cnn} hardware accelerator NE16 (based on RBE~\cite{contiRBE}) is specialized for highly efficient MAC operations. NE16 features 9x9x16 8x1bit MAC units, which are optimally used in 3x3 convolutions, but it also offers support for 1x1 and 3x3 depth-wise convolutions and fully connected layers. GAP9 has a hierarchical memory layout, with \SI{128}{\kibi\byte} of high-bandwidth, single-cycle-accessible L1 scratchpad memory in the cluster, \SI{1.5}{\mebi\byte} of interleaved L2 memory for data and code as well as \SI{2}{\mebi\byte} of on-chip flash memory. GAP9 also offers a rich set of peripherals for connecting to external memory, sensors and standard interfaces such as UART.

\subsection{MCU Deployment}
Different \gls{mcu} platforms require different deployment tools, which we introduce in this section. Note that we deploy networks to two compute domains on GAP9 (the RISC-V cluster and the NE16 \gls{cnn} accelerator) using two different flows as described below.
\label{subsec:deployment}
\subsubsection{General-Purpose Processor Deployment}
For deployment to GAP9's cluster, we use the DORY framework~\cite{burrello2020toc}. DORY is an automated deployment utility for ultra-low-power edge platforms with hierarchical memory layouts. It takes a precision-annotated ONNX file as the input and generates ANSI C code which implements the specified network on the target platform. Tiling between up to three hierarchical memory levels (L1 scratchpad memory, L2 main on-chip memory and L3 off-chip memory) is automatically performed with an \gls{ilp}-based tiling algorithm which takes into account the hardware-specific constraints (i.e., the memory size of each hierarchical level) and various heuristics.

For deployment on the ARM Cortex-M4/M7 platforms, we use TensorFlow-Lite Micro~\cite{david_tensorflow_2021}. As such, the deployment on the single-core ARM Cortex-M4 and M7 is performed by quantizing the network weights to 8-bits, generating the C++ code using TensorFlow-Lite Micro, and compiling the code for the corresponding microcontroller. 

\subsubsection{CNN Accelerator Deployments}
The deployment on MAX78000's \gls{cnn} accelerator was performed by using Analog Devices' training framework called \emph{ai8x-training} \footnote{\url{https://github.com/MaximIntegratedAI/ai8x-training/}} and deployment framework called \emph{ai8x-synthesis} \footnote{\url{https://github.com/MaximIntegratedAI/ai8x-synthesis/}}. After \gls{qat} with ai8x-training, the ai8x-synthesis framework is used to quantize the network weights, activations and input using a forked version of the Neural Network Distiller by Intel AI Lab \footnote{\url{https://nervanasystems.github.io/distiller}}. Finally, the C code used to deploy the network on the MAX78000's \gls{cnn} accelerator is generated by the "izer" tool, which converts the quantized trained model into C code. The complete network fits the accelerator and all the weights can be stored inside the weight memory of the accelerator.

Deployment of the network on GAP9's NE16 neural engine was done in collaboration with Greenwaves Technologies. To deploy networks on the GAP9 microcontroller, Greenwaves Technologies distributes a deployment framework called NNTool as part of the GAP SDK \footnote{https://github.com/GreenWaves-Technologies/gap\_sdk}. NNTool is used for \begin{enumerate*}[label=(\roman*),,font=\itshape] \item post-training network quantization \item network evaluation for activation and parameter sizing and \item code generation for deployment.\end{enumerate*} Analogous to DORY, it calculates a tiling of the model's individual layers such that the data for each tile fits into the L1 scratchpad.

\subsubsection{End-to-End evaluation}
\begin{figure}[htbp]
\centerline{\includegraphics[width=\columnwidth]{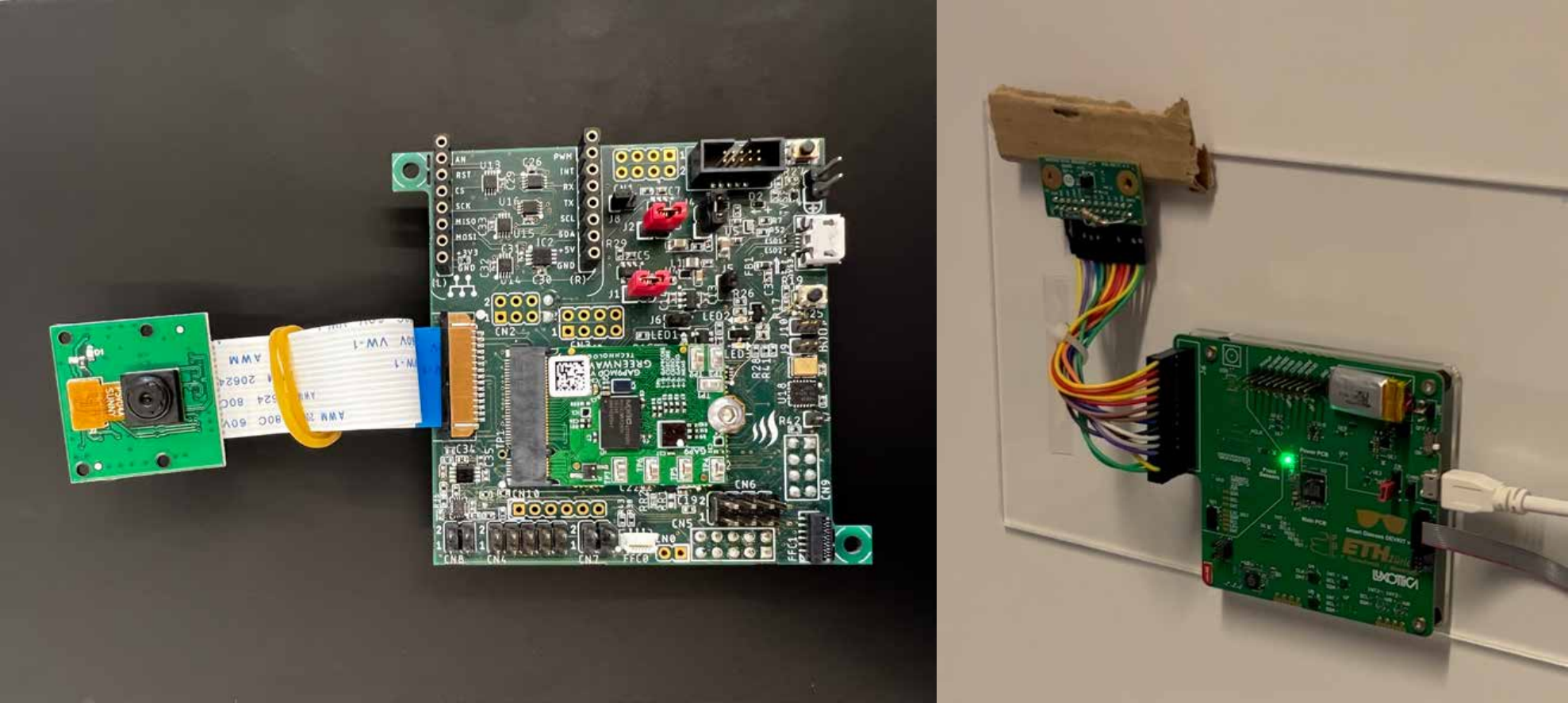}}
\caption{Left: GAP9 evaluation board with the OV5647 camera attached. Right: custom-developed PCB featuring the MAX78000 with the OVM7692 camera attached.}
\label{fig:demo_boards}
\end{figure}

To perform an accurate measurement including the image acquisition, we implemented the complete sensing pipeline consisting of the microcontroller platforms running TinyissimoYOLO and attached RGB cameras. In this way, the edge inference can be performed with real-world data, without relying on synthetic data from the dataset. We attached an OV5647 RGB CMOS camera from Omnivision to the GAP9 and attached the OVM7692 CameraCubeChip
to the MAX78000. As such we are able to demonstrate the functionality of the trained networks on two different devices. We performed our measurements on the GAP9 evaluation Kit and on a custom-designed PCB for the MAX78000, shown in \cref{fig:demo_boards}.

\begin{figure}[htbp]
    \centerline{\includegraphics[width=\columnwidth]{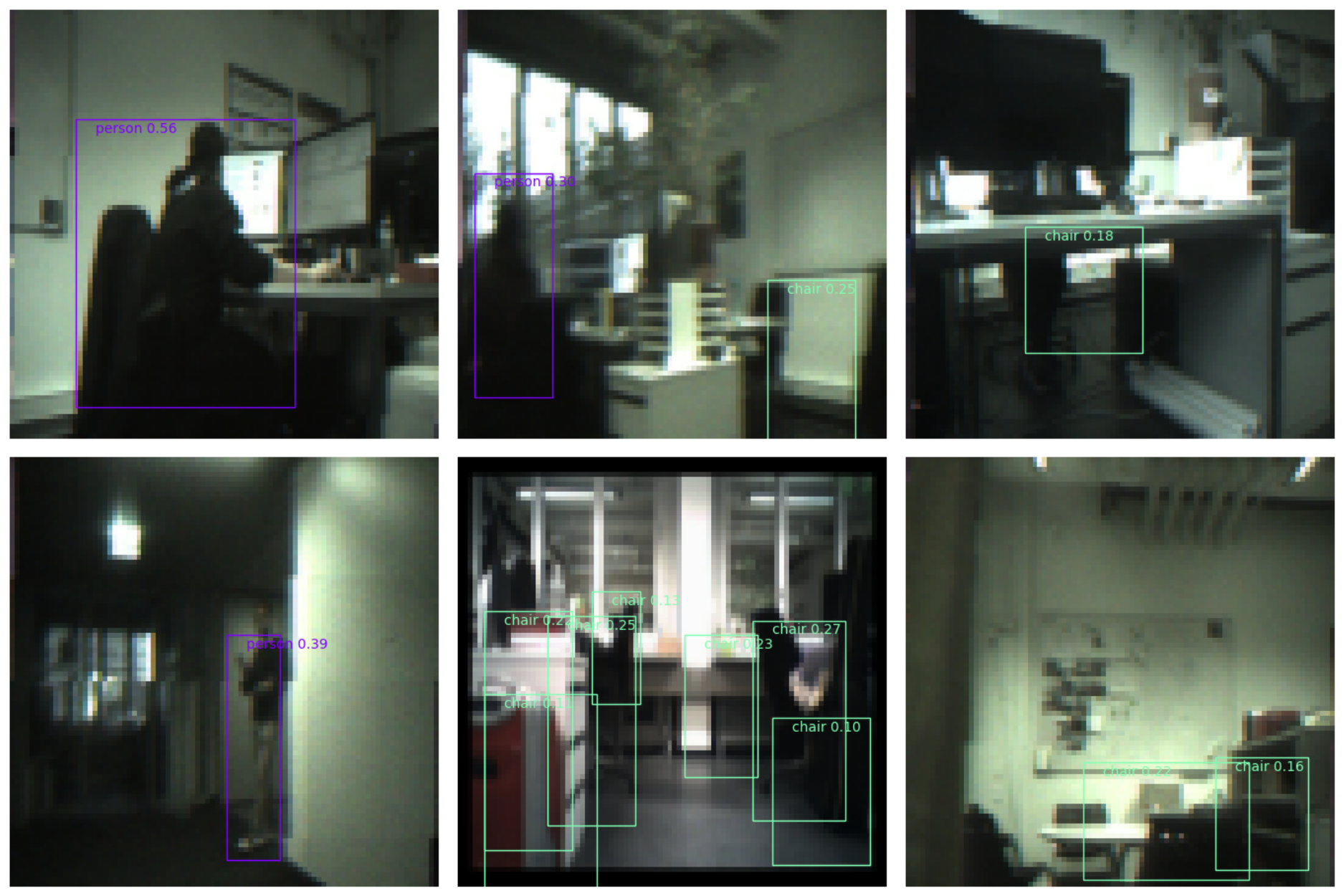}}
    \caption{Example recordings of images with GAP9 and running an inference of TinyissimoYOLO on the recorded images}
    \label{fig:demo_images_gap9}
\end{figure}

\begin{figure}
    \centerline{\includegraphics[width=\columnwidth]{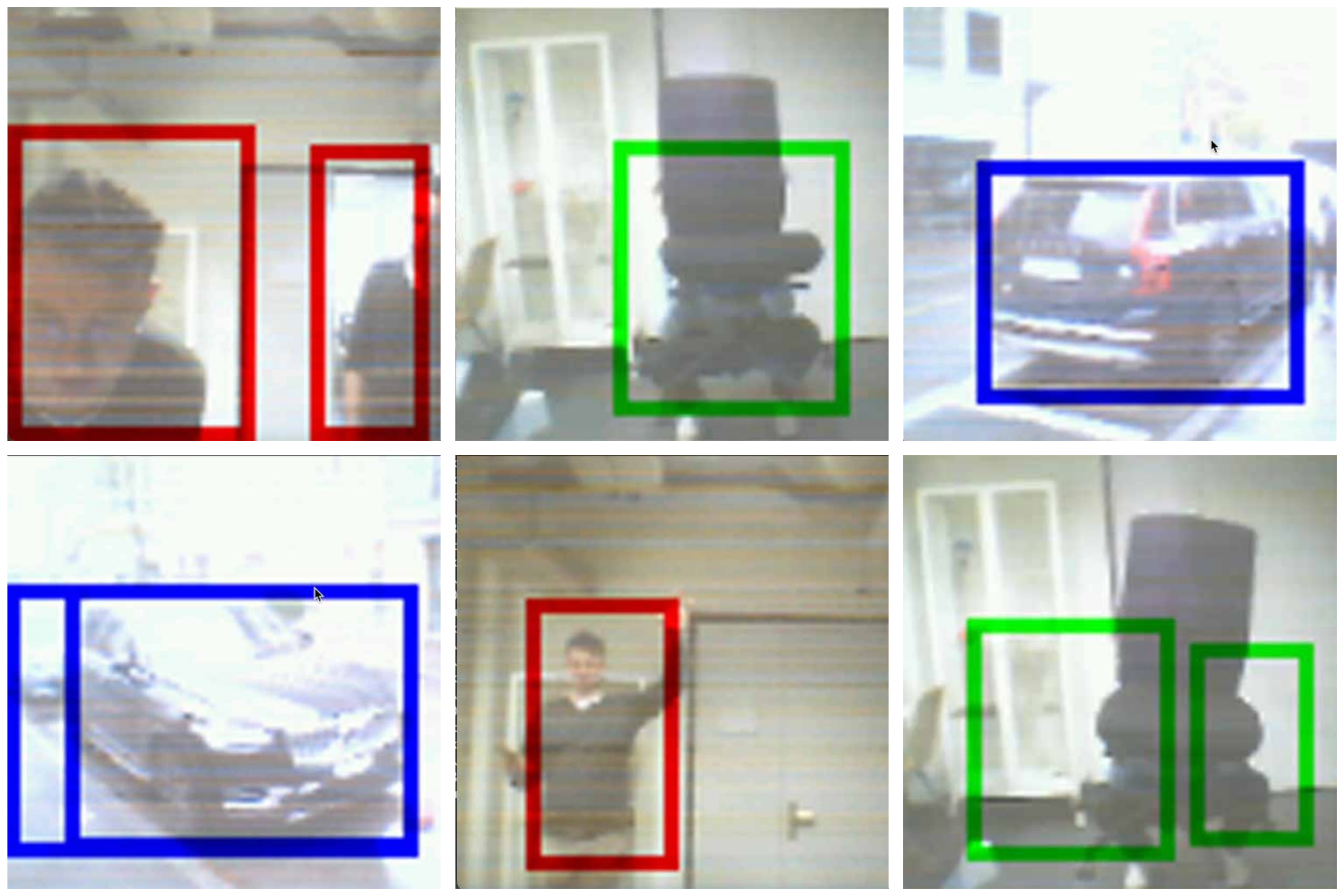}}
    \caption{Example recordings of images with MAX78000 running an inference of TinyissimoYOLO on the recorded images}
    \label{fig:demo_images_max78000}
\end{figure}

Further, \cref{fig:demo_images_gap9} shows some examples of images captured on GAP9 after processing with the TinyissimoYOLO network, while \cref{fig:demo_images_max78000} shows some other example images of a demo running image capturing and inference on the MAX78000, while streaming the image (after adding the detected boxes on the MAX78000 itself) live, via UART, to an attached PC.
\section{Experimental Results}
\label{sec:results}
To evaluate the trade-off between detection performance and energy efficiency, we trained fifteen different network variants on the PascalVOC dataset and report the \gls{map} for each class in \Cref{tab:map_tinyissimoyolo}. 
All network versions shown can be deployed on general purpose \glspl{mcu} such as GAP9. To compare the performance of GAP9 with the reported performance on the paper ~\cite{moosmann2023tinyissimoyolo}, we deployed TinyissimoYOLO (TY:3-3-88) on GAP9. First, we deployed it on single-core \glspl{mcu} only\footnote{Despite GAP9 being multi-core, the network deployment was done such that the inference runs only on one of the eight compute cores} to fairly compare to other single-core implementations such as STM32H7A3 featuring an ARM Cortex-M7, STM32L4R9 with an ARM Cortex-M4 and the sub-threshold computing microcontroller Apollo4b from Ambiq, also with a single-core ARM Cortex-M4. Then we also deployed a parallelized implementation, for the best possible performance on the eight-core RISC-V cluster of GAP9 without making use of the built-in neural engine hardware accelerator, showing the advantages of a parallel platform.
Using GAP9's capability to set the core voltage and frequency, we provide an analysis of different operating points, mapping a Pareto front of the latency-efficiency trade-off.
As a last step, the network is deployed on the NE16 neural engine of GAP9. \\
TinyissimoYOLO was limited by the \SI{442}{\kilo\byte} in accelerator weight-memory of the MAX78000, which restricted it to a network with only 3 of 20 available classes being trained, with an input resolution of $88\times88$ pixels to avoid relying on a specialized input streaming mode. As GAP9 features a bigger (\SI{1.5}{\mega\byte}) on-chip memory and can even use external memory, we also chose a second variant of the network, based on the results gathered. The second version features an input resolution of $112\times112$ pixels and is trained to detect 10 detection classes and has been deployed on GAP9 (TY:10-3-112) in a single-core, multi-core and neural engine accelerated version. We compare our implementations on GAP9 with the deployment on the MAX78000 \gls{mcu}. 
The single-core, multi-core, and neural engine deployment's performances are provided. We measured the power consumption of the whole \gls{soc} which we supplied with \SI{1.8}{\volt} and toggled a GPIO to detect the start and end of the \gls{cnn} execution.

\subsection{Network Architecture Variations}
\label{subsec:network_arch_variants_discussion}
The 15 trained networks vary in terms of network image input resolution, the first layer's kernel size, and the number of classes the network is trained for object detection. Notably, increasing the input resolution yields an increase of the input to the last fully connected layer, while increasing the number of classes yields an increase of the output of the last fully connected layer. In particular, by changing the input or the number of classes to detect, the number of network weight parameters will change accordingly. Therefore, \cref{tab:network_variation_overview} lists all the trained variants of TinyissimoYOLO and reports the number of parameters and the memory required to store all the network weights in quantized 8-bits.
\cref{tab:map_tinyissimoyolo} reports the \gls{map} for each detection class as well as the network's overall \gls{map}.
We varied the network input resolution between $88\times88$ pixels, $112\times112$ pixels, and $224\times224$ pixels. 
Even though we report training the network with an input resolution of $224\times224$ pixels, the network learning rate starting with 0.001 is unstable at the beginning leading the network to not get trained properly. However, by setting the initial learning rate to 0.0001, the network training is stable again, even though it takes more epochs to be trained. \\ When varying the number of object classes to predict, from 3 classes to 20 classes, for $88\times88$ pixel resolution the number of parameters scales by a factor of 1.3x, while for $224\times224$ pixel resolution, the scaling factor is 2x.
Notably, changing the network's first kernel to a kernel size of $3\times3$ to $7\times7$, we unsurprisingly note a constant increase of 1920 parameters.\\
Comparing the performance of the various TinyissimoYOLO networks, we first report the change of the first layer's kernel size from $3\times3$ to $7\times7$, does not increase the performance. In particular, it decreases overall performance constantly and rarely, for general poor performing classes, such as bottle or sofa, an increase of \gls{map} on a per class comparison can be reported, as for example TY:20-3-112 with 13\% \gls{map} for sofa while TY:20-7-112 with 16\% \gls{map}.
Increasing the input resolution, constantly increase the \gls{map} performance in overall network performance as well as in every class-to-class comparison. Notably, not surprisingly, decreasing the number of classes the network gets learned to detect increases the overall \gls{map} performance of the network and of each class itself. \\

\subsection{GAP9 - RISC-V MCU performance}
\begin{figure}[htbp]
\centerline{\includegraphics[width=\columnwidth]{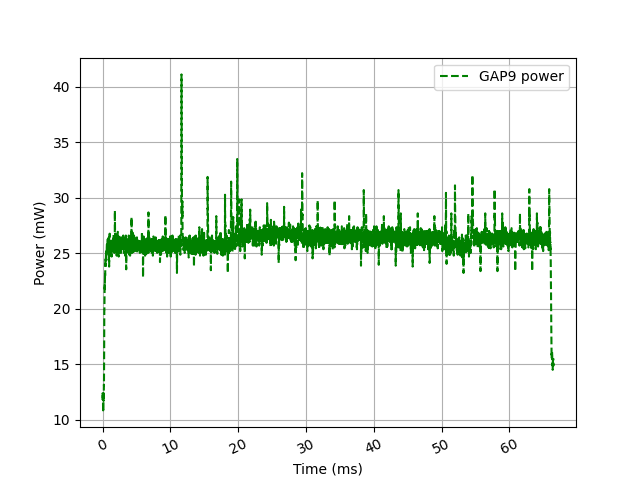}}
\caption{TY:3-3-88 single-core execution consumes \SI{26.14}{\mW} on average over \SI{69.77}{\ms}, resulting in an energy consumption of \SI{1738}{\uJ}.}
\label{fig:power_single}
\end{figure}

\begin{figure}[htbp]
\centerline{\includegraphics[width=\columnwidth]{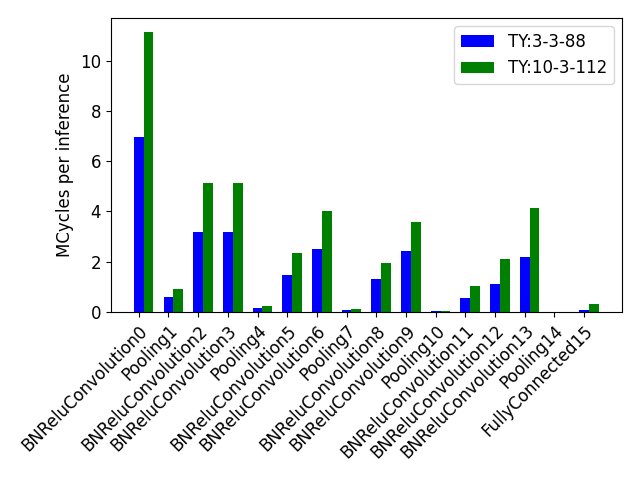}}
\caption{Single-core execution of TY:3-3-88 (in blue) achieves \SI{1.25}{MAC/cycle} and accumulates to a total of \num[group-separator={,}]{26} Mcycles, distributed to the different layers as shown here. Single-core execution of TY:10-3-112 (in green) achieves \SI{1.29}{MAC/cycle} and accumulates to a total of \num[group-separator={,}]{42} Mcycles, distributed to the different layers as shown here.}
\label{fig:cycles_single}
\end{figure}

We deployed two different \glspl{cnn}, the original TinyissimoYOLO (TY:3-3-88) for a fair comparison and an adapted network (TY:10-3-112) for more classes and higher accuracy. 
\paragraph{Single-core performance}
\label{sec:single_core}
Here we report the single-core performance of both networks deployed on GAP9.\\
\textbf{TY:3-3-88:}
Single-core execution on GAP9 results in \num[group-separator={,}]{26} Mcycles, so an equivalent of \SI{69.77}{\ms} at the maximum frequency of \SI{370}{\MHz} while we reach 1.25 MAC/cycles. The average power consumption is \SI{26.14}{\mW}, which gives us an energy consumption of \SI{1738}{\uJ}. In \cref{fig:power_single} we show the power consumption of the single-core implementation, showing a stable power consumption with a ripple, possibly from the internal voltage regulator as we measure the whole \gls{soc} power. In \cref{fig:cycles_single} we show the number of cycles by layer in blue - we see that the convolutional layers are the most computationally expensive, especially the first one. 

\begin{figure}[htbp]
\centerline{\includegraphics[width=\columnwidth]{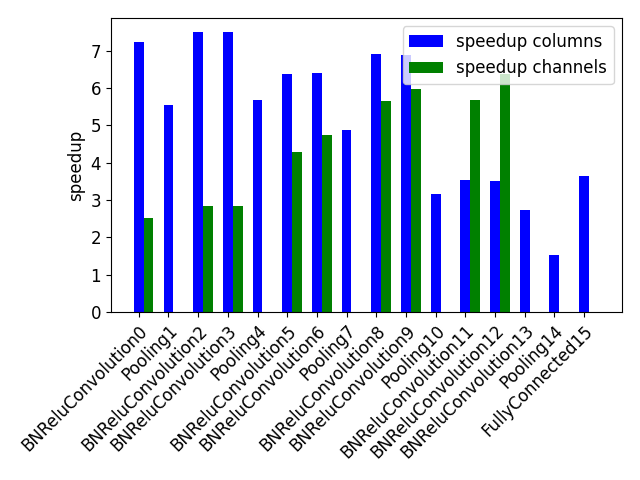}}
\caption{TY:3-3-88 speedup from 1 to 8 cores per layer for the two different parallelization schemes for convolutional layers.}
\label{fig:speedup}
\end{figure}

\textbf{112\texorpdfstring{$\times$}{x}112 input 10 classes:}
The single-core execution time of this network is \SI{114.15}{\ms} while consuming \SI{2990}{\micro\joule} per inference. We show the cycles per layer in \cref{fig:cycles_single} in green, seeing again that the first layer consumes most cycles and max-pooling is insignificant with only 3\% of the overall number of cycles.
\paragraph{GAP9 Multi-Core \gls{mcu} performance}
We deployed two different \glspl{cnn}, first the original TinyissimoYOLO for a fair comparison and then an adapted network for more classes and higher accuracy. Here we report the multi-core performance of both networks.\\
\textbf{TY:3-3-88:} In \cref{fig:speedup} we show the speedup for execution on 8 versus 1 core, for parallelizing by columns and, for the convolutional layers, by output channels.
We first parallelized by columns, which gives us good results for the first layers, however, leads to low speedup on small spatial dimensions of feature maps. Therefore we switched to parallelizing by output channels for layers 11 and 12, gaining 245k cycles. Layer 13 can not be parallelized by output channels, as it requires a higher stack size than what we can allocate. 

Overall we achieve a speedup of 6.14x running on 8 versus 1 core, only needing \num[group-separator={,}]{4.4} MCycles.

GAP9 can run at frequencies up to \SI{370}{\MHz} on both the fabric controller and the cluster while maintaining an inference efficiency of \SI{7.73}{MAC/cycle} on the multi-core cluster. We measured energy consumption and latency for different operating points between \SI{50}{\MHz} and \SI{370}{\MHz}, always choosing the minimum core voltage at which the system is still able to operate (in \SI{50}{\mV} steps). The MAC/cycle is not dependent on the frequency, meaning the latency scales linearly with the frequency.

\begin{figure}[htbp]
\centerline{\includegraphics[width=\columnwidth]{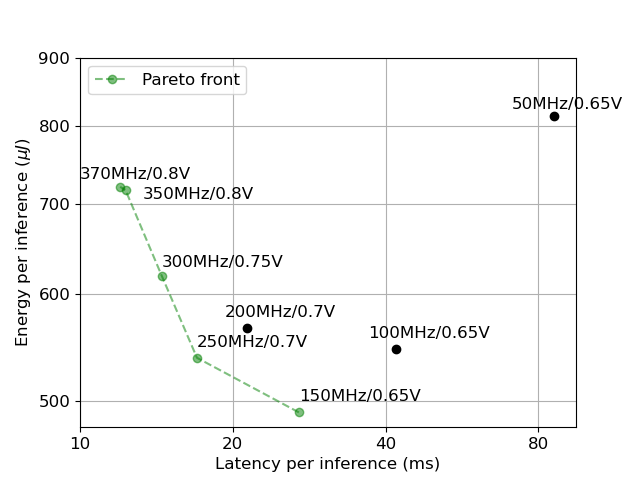}}
\caption{Latency versus energy efficiency at different operating points, showing the Pareto optimal set in green.}
\label{fig:pareto}
\end{figure}

In \cref{fig:pareto} we show our results, marking Pareto front points in green. We reach the most energy-efficient point at \SI{150}{\MHz}, as this allows operation at the minimum voltage of \SI{0.65}{\volt}. At this operating point, GAP9 only consumes \SI{490.21}{\uJ} per inference and exhibits a latency of \SI{27.9}{\ms}. 
Peak performance is reached at the maximum operating frequency of \SI{370}{\MHz}, with only \SI{11.3}{\ms} latency and an inference energy of \SI{721}{\uJ}.

\begin{figure}[htbp]
\centerline{\includegraphics[width=\columnwidth]{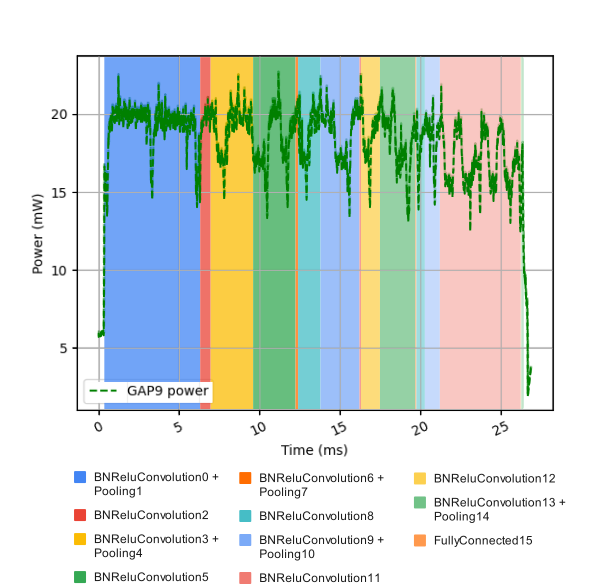}}
\caption{Energy-efficient TinyissimoYOLO on GAP9 has a latency of \SI{27.87}{\ms}  and an average power consumption of \SI{18.16}{\mW}.}
\label{fig:power_energy}
\end{figure}

\begin{figure}[htbp]
\centerline{\includegraphics[width=\columnwidth]{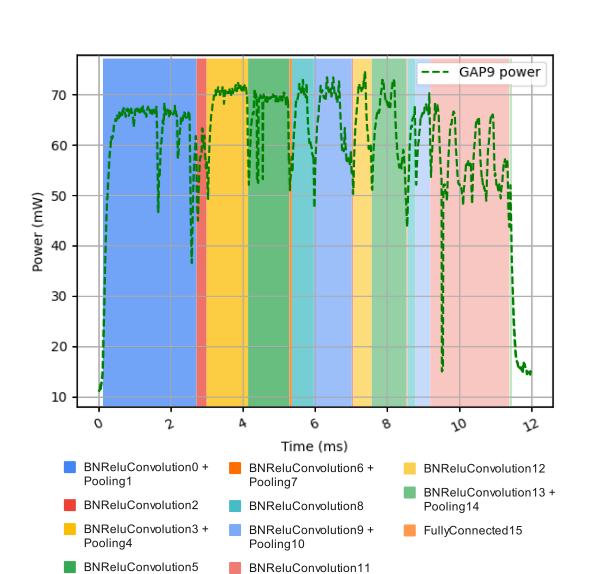}}
\caption{Peak performance TinyissimoYOLO on GAP9 has a latency of \SI{11.3}{\ms}  and an average power consumption of \SI{55.76}{\mW}.}
\label{fig:power_peak}
\end{figure}

In \cref{fig:power_energy}, we show the power consumption per layer for the most energy-efficient operating point (\SI{150}{\MHz} at \SI{0.65}{\volt}). We see the resulting latency, \SI{27.87}{\ms}, which corresponds to an energy per inference of \SI{490}{\uJ}.
In \cref{fig:power_peak} we show the power consumption per layer for the least latency operating point, the aforementioned \SI{370}{\MHz} at \SI{0.8}{\volt}. We see the resulting latency, \SI{11.3}{\ms}, which with the measured energy efficiency of \SI{162}{\uW/\MHz} corresponds to an energy per inference of \SI{721}{\uJ}.

\begin{figure}[htbp]
\centerline{\includegraphics[width=\columnwidth]{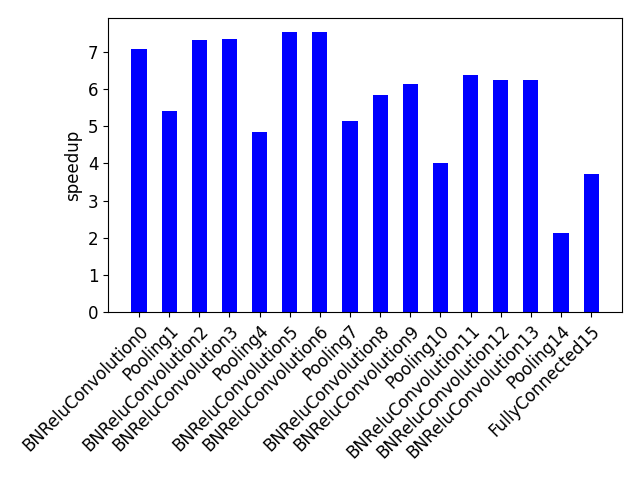}}
\caption{TY:10-3-112 speedup from 1 to 8 cores per layer, resulting in an average speedup of 6.77 and \SI{8.74}{MAC/cycle}.}
\label{fig:speedup_112}
\end{figure}

\begin{figure}[htbp!]
\centerline{\includegraphics[width=\columnwidth]{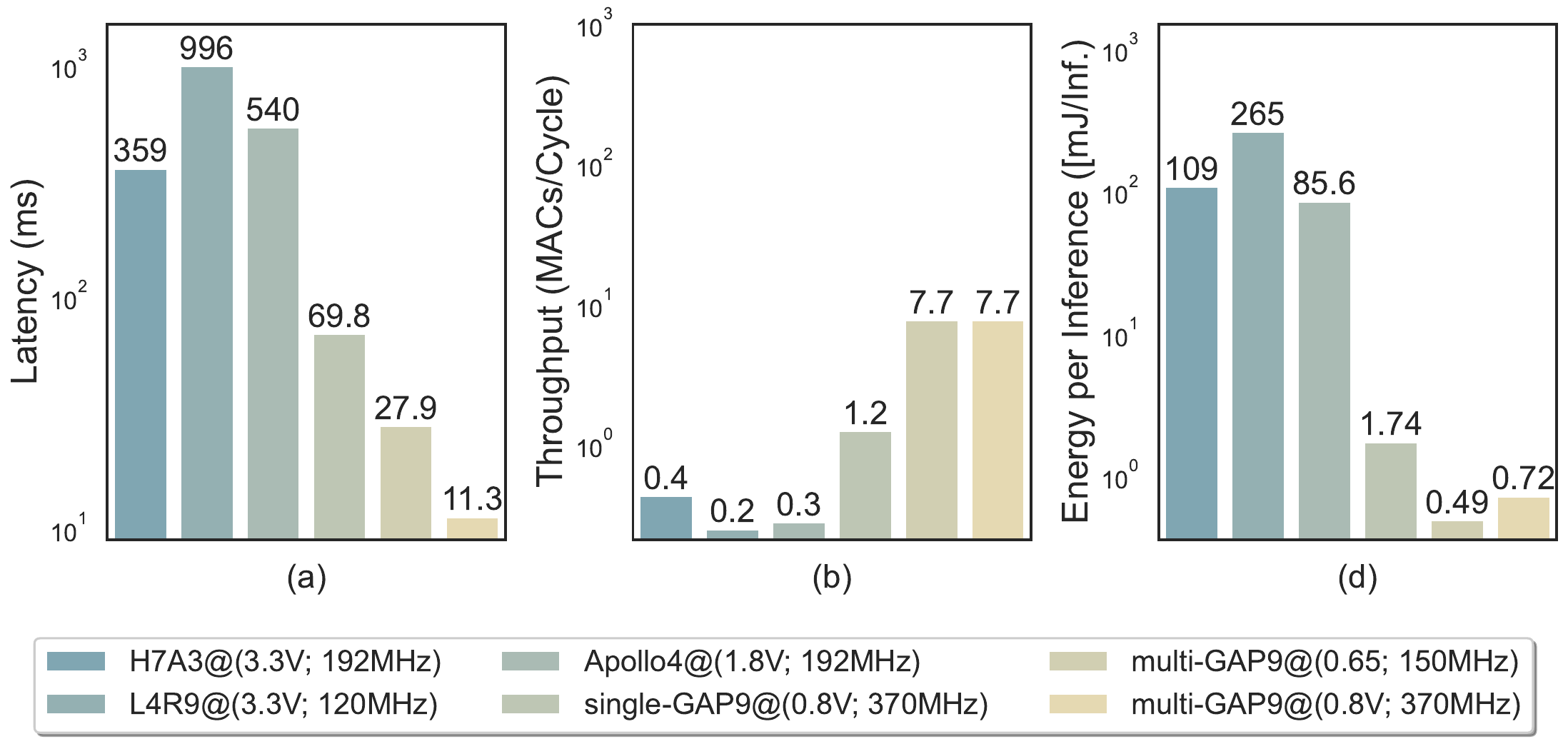}}
\caption{TY:3-3-88 performance comparison when deployed quantized to 8-bit on different \gls{mcu} architectures. The GAP9 single- and multi-core implementation outperform the other architectures in terms of latency, inference efficiency, and energy per inference.}
\label{fig:dev_comparison_single}
\end{figure}

In both \cref{fig:power_energy} and \cref{fig:power_peak} we visualize the execution times of the different layers by different colors. We can note a tendency that layers that parallelize better consume more power - which is expected, as the cluster then is fully used.
 \\
\textbf{TY:10-3-112:}
We achieve an overall speedup of 6.77 when parallelizing on 8 cores. In \cref{fig:speedup_112} we show the speedup per layer, as before earlier convolutional layers parallelize better due to the higher number of columns - however, in this network, there is no need to parallelize by output channels as the speedup of the last convolutional layers is still above 6. At peak performance (\SI{370}{\MHz}) we achieve a latency of \SI{16.87}{\ms} while consuming \SI{1057}{\uJ} per inference. Optimizing for energy-efficiency and running at \SI{150}{\MHz}, latency increases to \SI{41.62}{\ms} while consuming \SI{765}{\uJ}.
\paragraph{Comparison to ARM}
In \cref{fig:dev_comparison_single} we compare our deployment on GAP9 RISC-V cores (single-core as well as multi-core) against deployments on the H7A3, L4R9, and Apollo. Comparing the single-core versions, GAP9 clearly outperforms the other architectures in terms of latency (by a factor of more than 10 to the next best architecture) and energy per inference (by a factor of almost 50). In inference efficiency, we can also outperform the other architectures, as we can execute vectorized 8-bit operations on GAP9 contrary to the other architectures. 
For our multi-core implementation, we show the most energy efficient (\SI{150}{\MHz}) and least latency (\SI{370}{\MHz}) operating points of GAP9. Compared to the peak performance single-core implementation we can either achieve similar latency and around 3x reduced energy per inference or reduce latency by a factor of around 2.5 but only be around 2x more energy-efficient. 
\subsection{GAP9 Hardware Accelerator performance}
\begin{figure}[htbp]
\centerline{\includegraphics[width=\columnwidth]{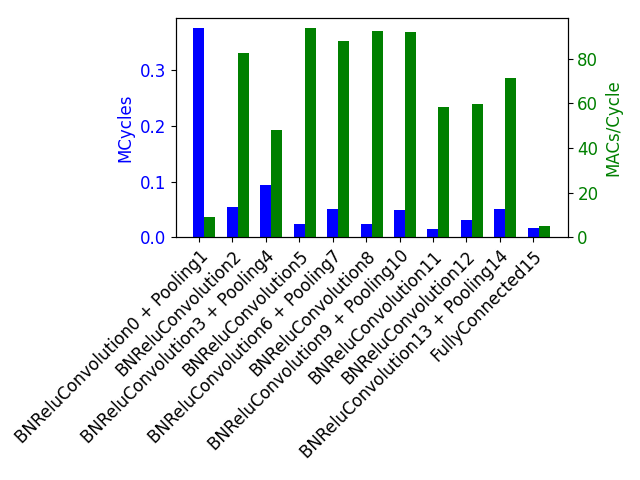}}
\caption{Using the HW accelerator on GAP9 we can reach an average of \SI{41.22}{MAC/cycle}, which leads to a total of \num[group-separator={,}]{785} kcycles.}
\label{fig:cycles_HW_88}
\end{figure}

\begin{figure}[htbp]
\centerline{\includegraphics[width=\columnwidth]{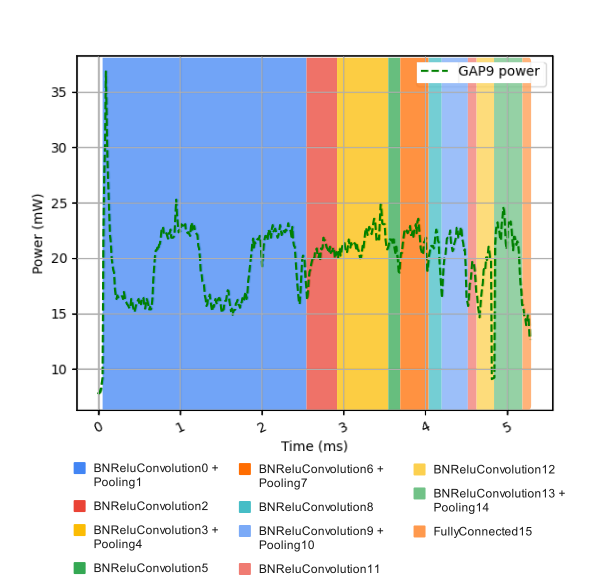}}
\caption{Energy-efficient TY:3-3-88 on the GAP9 hardware accelerator has a latency of \SI{5.24}{\ms} and an average power consumption of \SI{20.04}{\mW}.}
\label{fig:power_energy_HW}
\end{figure}

\begin{figure}[htbp]
\centerline{\includegraphics[width=\columnwidth]{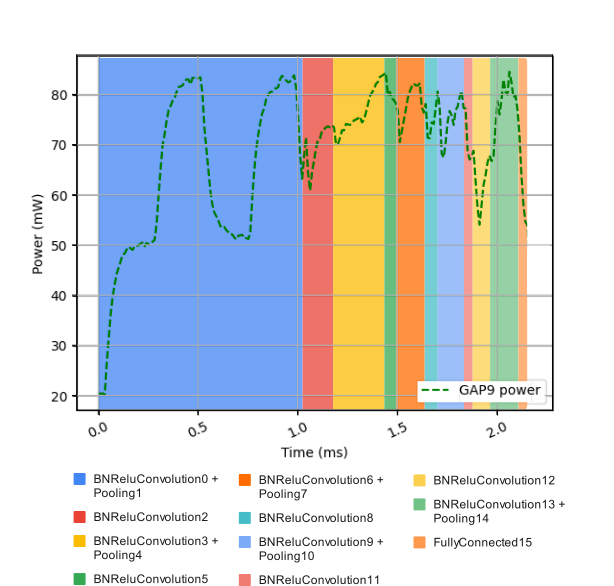}}
\caption{Peak performance TY:3-3-88 on the GAP9 hardware accelerator has a latency of \SI{2.12}{\ms}  and an average power consumption of \SI{70.30}{\mW}.}
\label{fig:power_peak_HW}
\end{figure}

\begin{figure}[htbp!]
\centerline{\includegraphics[width=\columnwidth]{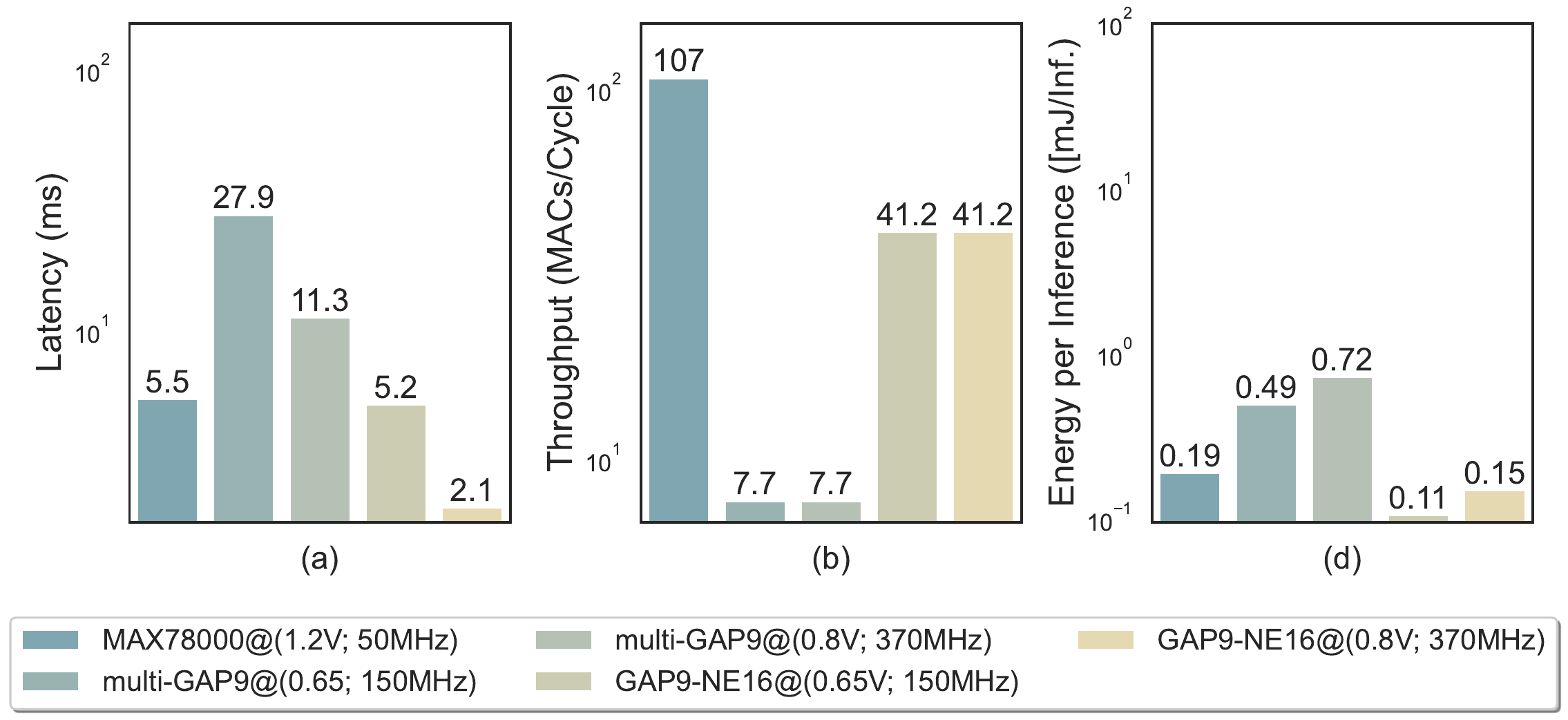}}
\caption{TY:3-3-88  performance comparison when deployed quantized to 8-bit on the \gls{cnn} accelerated MAX78000 (\SI{50}{\MHz}) \gls{mcu} compared to the most energy efficient (\SI{150}{\MHz}), the least latency (\SI{370}{\MHz}) operating points of GAP9 and the neural engine of GAP9.}.
\label{fig:dev_comparison_multi}
\end{figure}

We also deployed the proposed two networks on the Hardware accelerator on GAP9. \\
\textbf{TY:3-3-88:}
In ~\cref{fig:cycles_HW_88} we show the cycles and MAC/cycle per layer. We have an array of $9\times9$ NE16 engines that can handle 16 multiplications at a time, so are ideal for a multiple of 16 input channels. Note that max-pool layers can not be executed on the accelerator, but have to be computed on the cluster. Those two factors compromise the MAC/cycle number for the first layer. On average we achieved \SI{41.22}{MAC/cycle}.
In \cref{fig:power_energy_HW} (\SI{150}{\MHz}, the most energy-efficient operating point) and \cref{fig:power_peak_HW} (\SI{370}{\MHz}, the least latency operating point) we show the power consumption for the network running on the GAP9 hardware accelerator. We observe that while the hardware accelerator is active the power consumption is higher than when only the cluster is active (on max-pooling operations, which is executed tiled in two parts on the first layer).
At (\SI{370}{\MHz} we achieved a latency of \SI{2.12}{\ms} and an energy per inference of \SI{149}{\uJ}, which is a 5.3x speedup compared to only using the general purpose cores. At \SI{150}{\MHz} the latency is \SI{5.24}{\ms} while the energy per inference is \SI{105}{\uJ}, reducing the energy by 79\% compared to the multi-core implementation.
 \\
\textbf{TY:10-3-112:}
Deploying the $112\times112$ input network on the HW accelerator at \SI{370}{\MHz} we reached \SI{42.84}{MAC/cycle} and a latency of \SI{3.46}{\ms}, while consuming only \SI{245}{\uJ}. Compared to the multi-core implementation this is a 4.9x speedup. Running at \SI{150}{\MHz} we can improve the energy efficiency to \SI{177}{\uJ} per inference while increasing the latency to \SI{8.54}{\ms}, reducing the energy per inference by 77\% compared to the multi-core implementation.
\paragraph{Comparison to MAX78000}
In \cref{fig:dev_comparison_multi} we compare the most energy efficient (\SI{150}{\MHz}) and least latency (\SI{370}{\MHz}) operating points of GAP9 to the implementation on the \gls{cnn} accelerated MAX78000 \gls{mcu}. 
Experimental results show the MAX78000 outperforms the single-core and multi-core implementations on GAP9 in terms of latency, inference efficiency, and energy per inference.
However, the network implementation on the neural engine of GAP9 outperforms the inference latency and energy per inference of the MAX78000 because of the high clock frequency available. Even though the inference efficiency is 2.47x times less with \SI{43.2}{MAC/cycle}, GAP9 with the NE16 reaches a latency of only \SI{2.12}{\ms} and energy consumption of \SI{149}{\uJ} at peak performance, being 2.6x faster and 1.3x more energy efficient than the MAX78000. At the most energy efficient frequency GAP9 still reaches a slightly lower latency than the MAX78000, but can even reduce the energy consumption by a factor of 1.8.

\begin{figure}[htbp!]
\centerline{\includegraphics[width=\columnwidth]{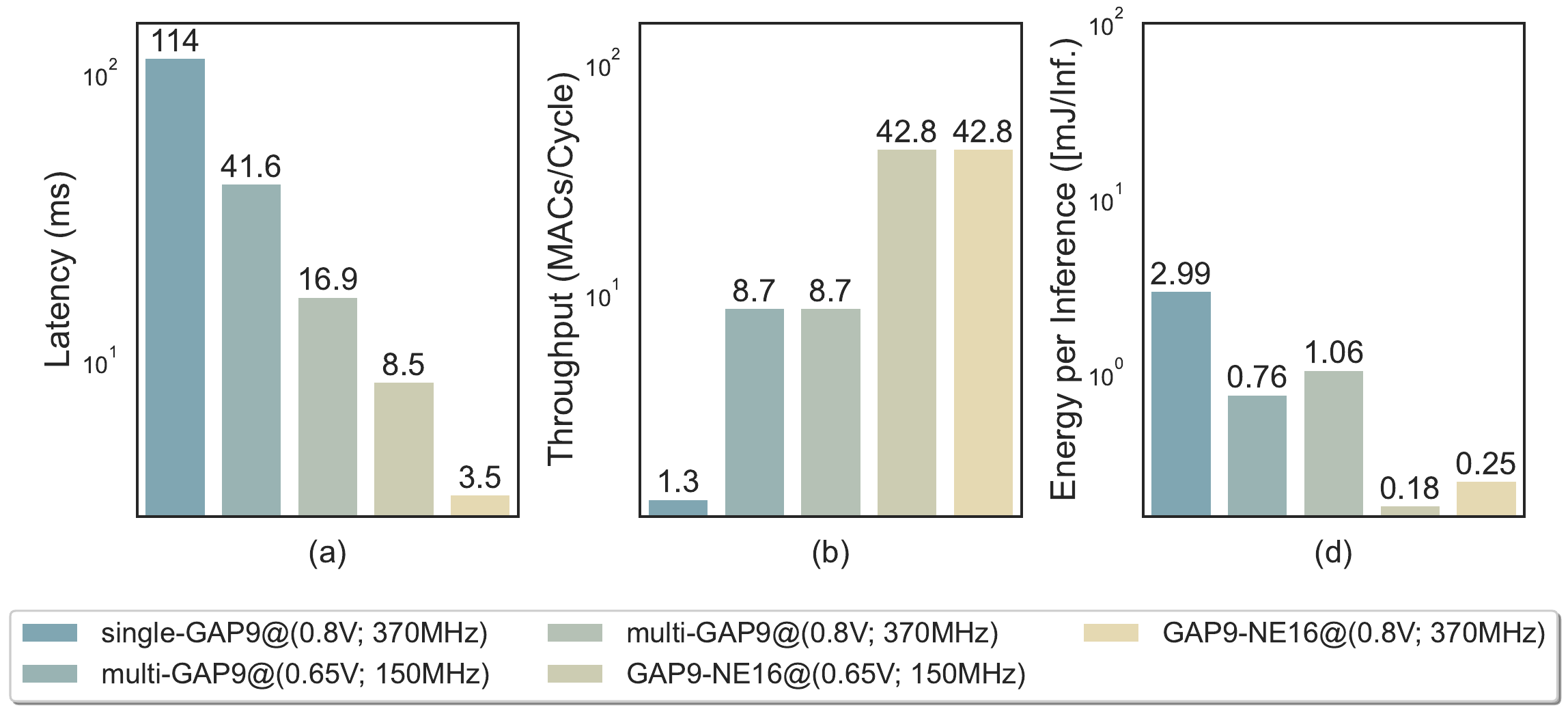}}
\caption{Performance comparison when deploying the network TY:10-3-112 quantized to 8-bit on the GAP9 running on the single-core (\SI{370}{\MHz}), multi-core (\SI{150}{\MHz}, \SI{370}{\MHz}) and on the neural engine of GAP9 (\SI{370}{\MHz})}.
\label{fig:dev_comparison_112gap9}
\end{figure}

By using the GAP9's multi-core processor or the neural engine, we gain flexibility, as it is a multi-purpose architecture that allows to deploy of arbitrary networks while the MAX78000 is limited to a specific set of layers and only internal memory. In section \cref{subsec:network_arch_variants_discussion} we showed that we can gain accuracy and train for more classes if we can use bigger networks that can be deployed on GAP9 but not on the MAX78000. 
\paragraph{Comparison TY:10-3-112:}
We also deployed a more general-purpose TinyissimoYOLO network on GAP9, which has a higher input resolution and 10 detection classes. In \cref{tab:network_variation_overview} we note the network has 700k parameters. Furthermore, the input of $112\times112$ pixels RGB image consumes another \SI{100}{\kilo\byte} of memory while the in-between network calculations need at most approximately \SI{375}{\kilo\byte} of memory. As such, a microcontroller that needs to run such a network requires at least \SI{1}{\mega\byte} of Flash while having \SI{512}{\kilo\byte} of RAM. Furthermore, this network clearly can not fit the MAX78000 anymore. We, therefore, deployed the network on GAP9 only. In particular, we deployed the network on one single core, on all eight cores and on the neural engine itself. \cref{fig:dev_comparison_112gap9} shows the comparison of the performances achieved. We note, despite the sheer size of the network, it runs within \SI{3.5}{\ms} on the neural engine, while being executed 32x and 3x slower on the single-core and multi-core implementation, respectively. 
\section{Conclusion}
\label{sec:conclusion}
This work provides a comprehensive evaluation of various network adjustments for TinyissimoYOLO for edge processors with a 100s-of-\si{\kibi\byte} memory budget and in a 10s-of-milliwatt-range power envelope. We demonstrate the versatility of TinyissimoYOLO by training the network to detect up to 20 classes. Despite its small size, TinyissimoYOLO achieves remarkably high detection accuracy, coming close to the performance of YOLOv1 when trained on the entire PascalVOC dataset. Particularly noteworthy is the fact that with an input resolution of $112\times112$, TinyissimoYOLO outperforms ten individual classes of the original YOLOv1 network within the 20 classes network.
Additionally, we present an exhaustive investigation into the network's deployability with a fair benchmark and discussion of single-core microcontrollers and the benefit of parallelization. Furthermore, the novel RISC-V-based multi-core GAP9 processor is compared with the MAX78000 accelerator and the GAP9's neural engine (NE16). On NE16, we find that inference at the maximum clock frequency of \SI{370}{\MHz} only takes \SI{2.12}{\ms}, with an energy consumption per inference of \SI{105}{\uJ}, nearly half the energy consumption compared to the MAX78000 platform for a 3-class network with $88\times 88$ input resolution. This network's architecture and size are largely dictated by the MAX78000 accelerator's limitations, which do not apply to GAP9's heterogeneous architecture - layers not supported by NE16 can still be efficiently mapped to the multi-core RISC-V cluster. We conclude that multi-core, general-purpose platforms are essential to achieving acceptable performance and efficiency levels. Heterogeneous systems incorporating domain-specific accelerators provide an efficiency boost in the accelerated applications, but the presence of tightly coupled general-purpose processors is essential to maintain flexibility. \\
This is illustrated by the deployment of a larger, more powerful 10-class TinyissimoYOLO network using a larger input resolution of $112\times112$ pixels to GAP9's cluster and to NE16. Even with this more powerful network, GAP9 can perform 285 inferences per second, with the object detection update rate ultimately restricted by the exposure time of low-power cameras rather than inference latency. The remaining time between frames could be used for additional processing of the inference results, e.g. for object tracking. In conclusion, the energy efficiency and real-time capabilities of TinyissimoYOLO make it well-suited for low-power processors and applications such as always-on smart cameras, where it can perform object detection efficiently. 

\begin{acks}
To Marco Fariselli, for his help in deploying TinyissimoYOLO on the neural engine of GAP9, to Thorir Mar Ingolfsson, for his help with DORY on GAP9, and to Jakub Mandula for proofreading this paper.

Further, the authors would like to thank armasuisse Science \& Technology for funding this research.
\end{acks}




\bibliographystyle{ACM-Reference-Format}
\bibliography{sample-base}

\appendix









\end{document}